\documentclass{article}

\usepackage[main, final]{neurips_2026}

\usepackage[utf8]{inputenc}
\usepackage[T1]{fontenc}
\usepackage{hyperref}
\usepackage{url}
\usepackage{booktabs}
\usepackage{amsfonts}
\usepackage{nicefrac}
\usepackage{microtype}
\usepackage{xcolor}

\newcommand{\name}{{MAGE}\xspace}
\newcommand{\mask}{\texttt{[MASK]}}
\newcommand{\fullname}{{MAGE}\xspace}

\usepackage{xspace}
\usepackage{listings}
\usepackage{placeins}
\usepackage{graphicx}

\newcommand{\para}[1]{\noindent \textbf{#1.}}

\newcommand{\Scal}{\mathcal{S}}

\usepackage{amssymb}

\usepackage{amsmath}

\usepackage{mathtools}

\usepackage{multirow}
\usepackage{threeparttable}

\usepackage{algorithm}
\usepackage{algpseudocode}
\usepackage{lipsum}
\usepackage{enumitem}
\usepackage{tikz}
\DeclareRobustCommand{\bcircle}[1]{%
  \tikz[baseline=-0.6ex]{
    \node[circle, fill=black, text=white, minimum size=1.0em, inner sep=0.6pt] {\scriptsize\bfseries #1};
  }%
}

\title{\fullname: All-\mask Block Already Knows Where to Look in Block Diffusion LLM}

\author{%
  Omin Kwon \\
  Seoul National University \\
  \texttt{om0127@snu.ac.kr} \\
  \And
  Yeonjae Kim \\
  Seoul National University \\
  \texttt{duswo1120@snu.ac.kr} \\
  \And
  Doyeon Kim \\
  Seoul National University \\
  \texttt{pega501@snu.ac.kr} \\
  \AND
  Minseo Kim \\
  UC Berkeley \\
  \texttt{minseo.kim@berkeley.edu} \\
  \And
  Yeonhong Park \\
  Meta \\
  \texttt{parkyh96@gmail.com} \\
  \And
  Jae W. Lee\thanks{Corresponding author.} \\
  Seoul National University \\
  \texttt{jaewlee@snu.ac.kr} \\
}

\begin{document}

\maketitle

\begin{abstract}
Block diffusion LLMs are an emerging paradigm for parallel language generation, but their KV caching makes memory access the dominant bottleneck in long-context inference.
Sparse attention, which attends only to a small KV subset per query, can reduce this latency with minimal accuracy loss.
In block diffusion, however, the $B$ tokens of each block must share a single KV subset, and we show this per-block constraint degrades existing sparse KV estimators by up to $25\%$ in recall.
We address this challenge by exploiting a property that emerges from the block-diffusion training objective: it aligns the block-average query across denoising steps, so the All-\texttt{[MASK]} block at the first step already reveals the per-block KV subset for the entire trajectory.
We exploit this in \name (\texttt{[MASK]}-Guided Sparse Attention), a training-free method that runs one exact attention pass at the first step and reuses its top-$k$ index sets for all remaining steps within the block.
Across three block-diffusion families on LongBench, \name matches Exact Attention at $k{=}512$ with near-lossless accuracy, achieves up to $6.82\times$ end-to-end speedup at $128$K context, and runs up to $3.35\times$ and $2.28\times$ faster than Quest and SparseD, designed for AR LLMs and fully bidirectional diffusion LLMs, respectively.
The code is available at \url{https://github.com/SNU-ARC/MAGE}.
\end{abstract}

\section{Introduction}

Block diffusion LLMs (Block dLLMs)~\citep{arriola2025blockdiffusioninterpolatingautoregressive, wu2025fastdllmv2efficientblockdiffusion, cheng2025sdarsynergisticdiffusionautoregressionparadigm, bie2025llada20scalingdiffusionlanguage} have emerged as a promising paradigm for parallel language generation: by adapting pretrained autoregressive (AR) models into block-wise diffusion models with exact KV caching, they increase the throughput by multi-token generation while maintaining competitive quality.
As context length grows, attention dominates the overall inference computation and becomes bottlenecked by KV cache memory access~\citep{pope2022efficient}, motivating sparse attention as a natural remedy.
Sparse attention reduces this cost by attending only to a small selected \emph{KV subset}---the top-$k$ KV cache entries that capture most of the attention mass---rather than the full cache.
Dynamic estimators such as Quest~\citep{tang2024questqueryawaresparsityefficient} have proven highly effective for AR LLMs, and methods such as SparseD~\citep{wang2025sparsedsparseattentiondiffusion} have been proposed for fully bidirectional dLLMs---but neither was designed for block diffusion, and transferring them directly reveals a structural mismatch that substantially degrades their accuracy.

The mismatch stems from a constraint unique to block diffusion: the $B$ tokens within a block are forwarded together, so they must share a single \emph{per-block} KV subset rather than each selecting its own per-token subset.
Constructing a single shared subset that simultaneously satisfies each of the $B$ tokens' top-$k$ preferences is structurally harder: the union of their per-token subsets averages $4.5\times$ the budget. As a result, when existing estimators are adapted to this regime, their recall drops by up to $25\%$.
We identify an opportunity on a \emph{different axis}: block-diffusion training, by requiring the model to predict each masked token under all surrounding noise levels within a block, aligns the block-average query across the denoising steps.
As a consequence, the All-\texttt{[MASK]} block in the first step within a block already reveals the per-block KV subset for the entire trajectory, with per-block top-$k$ recall averaging over $84\%$ at $k{=}512$---substantially above the AR pre-training backbone, confirming this anchoring is training-induced.

\begin{figure}[t]
    \centering
    \includegraphics[width=\textwidth]{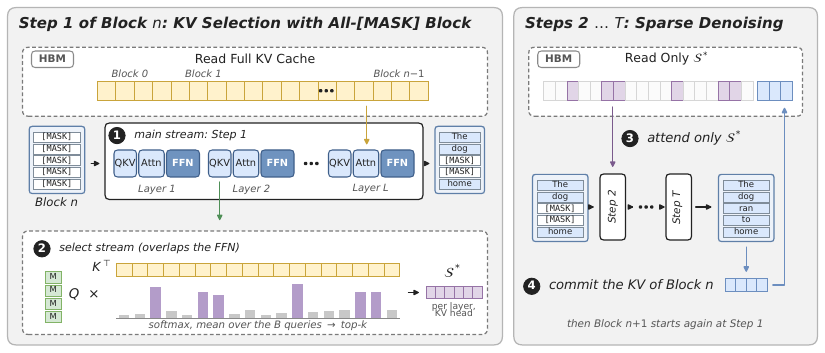}
    \caption{Overview of \name. \bcircle{1}--\bcircle{2}~At Step~1 of each block, whose $B$ positions are all \texttt{[MASK]}, the main stream runs the full forward pass while the select stream takes the queries of each layer, computes their attention over the prefix KV cache, averages it over the $B$ queries, and keeps the top-$k$ KV subset $\mathcal{S}^*$ per layer and KV head, overlapping the FFN. \bcircle{3}~Steps~2 to $T$ attend only to $\mathcal{S}^*$. \bcircle{4}~Once the block is denoised, its KV is computed and appended to the cache, and the next block starts again at Step~1.}
    \label{fig:overview}
\end{figure}

Building on this observation, we propose \name (\texttt{[MASK]}-Guided Sparse Attention), the first sparse attention method tailored to block-diffusion LLMs (Fig.~\ref{fig:overview}).
\name runs a single exact attention pass on the All-\texttt{[MASK]} block at the first step (Step 1), extracts the exact top-$k$ index sets, and reuses them unchanged for all $T{-}1$ remaining denoising steps---requiring no re-estimation at any subsequent step.
To avoid paying the selection cost on the critical path, \name executes the index-selection kernels asynchronously on a dedicated CUDA stream, overlapping them with the compute-intensive FFN of the main stream; the amortized non-sparse overhead is $1/T$ of one exact attention pass per step.
Across three block-diffusion LLM families on LongBench, \name matches Exact Attention at $k{=}512$ with near-lossless accuracy and achieves end-to-end speedups of up to $6.82\times$ at $128$K context, running up to $3.35\times$ faster than Quest and $2.28\times$ faster than SparseD.

Our contributions are summarized as follows:
\begin{itemize}[left=0.5cm]
    \item We identify the \emph{per-block shared KV selection} challenge unique to block diffusion, showing that existing sparse KV estimators lose up to $25\%$ recall under this constraint (\S\ref{subsec:challenge}).
    \item We discover the All-[\texttt{MASK}] anchoring property induced by block-diffusion training, and explain through a quantitatively verified five-link chain why Step 1 serves as an accurate anchor for the full denoising trajectory (\S\ref{subsec:allmask-knows}).
    \item We propose \name, a training-free sparse attention method with an asynchronous dual-stream design that incurs only $+1.9\%$ to $+6.4\%$ latency overhead at Step $1$ (\S\ref{sec:method}).
    \item We demonstrate near-lossless accuracy at $k{=}512$ with up to $6.82\times$ end-to-end speedup at $128$K context, outperforming Quest by up to $3.35\times$ and SparseD by up to $2.28\times$ (\S\ref{sec:experiments}).
\end{itemize}

\section{Related Work}

\subsection{Evolution of Diffusion Language Models}

Diffusion LLMs generate text by iteratively denoising sequences of \texttt{[MASK]} tokens, enabling parallel generation of multiple tokens per forward pass and breaking the left-to-right dependency of autoregressive (AR) models~\citep{nie2025large, ye2025dream7b}.

\para{Fully Bidirectional Diffusion LLMs}
Earlier diffusion LLMs~\citep{nie2025large, ye2025dream7b} apply bidirectional attention over the entire sequence, requiring the entire sequence to be recomputed at every denoising step.
This design is incompatible with standard KV caching and leads to substantial inference latency~\citep{kim2025beyond}, and also exhibits suboptimal task performance compared to strong AR baselines~\citep{nie2025large, ye2025dream7b}.
While subsequent works have proposed approximate KV caching schemes to partially mitigate this inefficiency~\citep{wu2025fastdllmtrainingfreeaccelerationdiffusion, ma2025dkvcachecachediffusionlanguage, dyllm}, such approximations inevitably introduce errors that accumulate across denoising steps.

\para{Block Diffusion LLMs}
More recent work departs from bidirectional architectures and instead trains models with block-wise generation in mind~\citep{arriola2025blockdiffusioninterpolatingautoregressive}.
Some approaches convert bidirectional diffusion models into block diffusion models~\citep{wang2025d2f, kim2026cdlmconsistencydiffusionlanguage}, while the more prevalent approach adapts pretrained AR models into block diffusion models~\citep{gong2025scalingdiffusion}, as exemplified by Fast-dLLM v2~\citep{wu2025fastdllmv2efficientblockdiffusion}, SDAR~\citep{cheng2025sdarsynergisticdiffusionautoregressionparadigm}, LLaDA 2.0~\citep{bie2025llada20scalingdiffusionlanguage}, TraDo~\citep{trado}, DreamReasoner~\citep{wu2026dreamreasoner8bblocksizecurriculumlearning}, and Nemotron-Labs-Diffusion~\citep{fu2026nemotronlabsdiffusiontrimodelanguagemodel}, and block diffusion has also been adopted by industry-developed models~\citep{gemini_diffusion, song2025seeddiffusionlargescalediffusion}.
These models enable \emph{exact}, block-level KV caching without approximation, shifting the primary bottleneck of attention from computation to KV cache memory access.
As a result, the efficiency of block diffusion LLM inference is increasingly dominated by KV cache memory access, particularly in long-context scenarios where the cache size grows substantially.
In this work, we focus on Block Diffusion LLMs and directly target this memory bottleneck.

\subsection{Sparse Attention for Long-Context Inference}
Sparse attention alleviates this bottleneck by accessing only the small subset of KV entries on which the attention mass concentrates---ideally the \emph{oracle top-$k$}, defined as the $k$ prefix indices that carry the bulk of the mass in the exact attention score distribution.
Obtaining the oracle, however, requires the very score matrix that exact attention over the full KV cache would produce; existing methods therefore perform sparse attention through \emph{lightweight estimators} of the oracle.
We measure how well any KV subset serves a given token via \emph{estimation accuracy}---the fraction of that token's oracle top-$k$ entries that are covered by the selected subset:
\begin{equation}
\mathrm{EstAcc}_{q,h}^{(t)} \;=\; \frac{\bigl|\mathcal{S}_{h}^{(t)} \cap \mathcal{S}_{q,h}^{*(t)}\bigr|}{k},
\label{eq:est-acc}
\end{equation}
where $h$ indexes the KV head, $\mathcal{S}_{h}^{(t)}$ is the KV subset being evaluated, and $\mathcal{S}_{q,h}^{*(t)}$ is the oracle KV subset for token $q$.
No such estimator has yet been designed for block dLLMs; below, we review existing methods developed for AR LLMs and fully bidirectional dLLMs, on which our analysis later builds.

\para{Methods for AR LLMs}
In AR LLMs, the KV cache grows token by token across decoding steps, and sparse attention has been studied most extensively in this setting.
Early AR sparse attention methods rely on fixed sparsity patterns---such as sliding windows or attention sinks where the attention mass concentrates~\citep{xiao2024streamingllm, beltagy2020longformer, zaheer2020bigbird}---or evict less critical KV entries based on accumulated attention statistics~\citep{zhang2023h2o, liu2023scissorhands, li2024snapkv, feng2025adakv}.
Neither fixed patterns nor one-shot eviction, however, can accurately track the oracle's preferred KV entries, which vary with every newly generated token at each decoding iteration.
This motivates \emph{dynamic} estimators that select a per-token KV subset on the fly~\citep{tang2024questqueryawaresparsityefficient, ribar2024sparq, singhania2024loki, yang2025tidaldecode, yang2025lessismore}.

Quest~\cite{tang2024questqueryawaresparsityefficient} is a representative example: it partitions the prefix KV cache into contiguous \emph{pages} of size $p$, summarizes each page by its per-channel max/min keys, scores each page against the query via a sign-aware upper bound on $q^{\!\top} k$ ($\sum_d \max(q_d K^{\max}_d, q_d K^{\min}_d)$), and uses the KV caches of the top-scoring pages as the sparse subset.
While page-summary estimation incurs some accuracy loss relative to the oracle, the ability to dynamically fetch a per-token KV subset substantially improves accuracy over fixed patterns.
Smaller $p$ yields finer-grained selection that tracks the oracle more closely, but the page-scoring step itself becomes more expensive---a granularity-cost trade-off.
Subsequent works refine this recipe in two directions: SparQ~\citep{ribar2024sparq} and Loki~\citep{singhania2024loki} reduce scoring cost via channel selection or low-rank projection, while PyramidKV~\citep{cai2024pyramidkv}, Tactic~\citep{zhu2025tactic}, and Twilight~\citep{lin2025twilight} adaptively tune the per-layer/per-head selection budget.

\para{Methods for Fully Bidirectional Diffusion LLMs}
In fully bidirectional dLLMs, the entire sequence's keys and values are recomputed at every denoising step without KV cache.
SparseD~\citep{wang2025sparsedsparseattentiondiffusion} performs exact attention during the first 20\% of the total denoising steps across all blocks, extracts each token's oracle top-$k$ KV subset at the final exact-attention step, and reuses these per-token attention masks across the remaining 80\% of the steps.
MaskKV~\citep{huang2025maskkv} instead drops less critical prompt tokens after prefill, reducing the sequence length processed in subsequent denoising steps.
In both methods, the sparse pattern is captured once per inference and reused throughout without further refresh as denoising progresses, which can potentially degrade recall as the trajectory unfolds.
Sparse-dLLM~\citep{song2025sparsedllmacceleratingdiffusionllms} instead evicts low-saliency KV entries with a delay, but an evicted entry can no longer be attended to.

\section{Challenges and Opportunities of Sparse Attention in Block Diffusion LLMs}
\label{sec:allmask}

In this section, we identify the structural constraint that sparse attention on block diffusion LLMs must operate under, and a block-diffusion-specific property that recovers estimation accuracy within this constraint along a different axis.

\subsection{The Challenge: Sparse KV Subset Estimation Is Harder in Block Diffusion LLMs}
\label{subsec:challenge}

Adapting sparse attention to block diffusion LLMs is structurally harder than it appears: preserving inference efficiency requires all block tokens to share a single KV subset via per-block shared memory access, and existing lightweight estimators suffer an estimation accuracy drop under this constraint.
We now explain why this conflict arises and quantify its impact.

\para{Per-Block KV Subset Forced by Shared Memory Access}
Within a block, the $B$ tokens pass through the model in a single model forward pass: their queries share the same prefix KV cache reads which include input prompts and the previously generated block, and the attention is computed as a single GEMM.
Therefore, unlike AR LLMs and fully bidirectional dLLMs---where each token selects its own \emph{per-token} sparse KV subset---a block dLLM requires the entire block to share a single \emph{per-block} sparse KV subset. Letting each token use its own subset is possible but makes the KV accesses irregular, which slows sparse attention by $2.5$--$7.9\times$ (Appendix~\ref{app:per-token}).
More precisely, the per-token KV subset is already shared across the $G$ query heads of a KV group under GQA~\citep{ainslie2023gqa}, and the block structure of block dLLMs introduces a further sharing across the $B$ token positions of the block, so the per-block top-$k$ averages scores over both the $G$ heads and the $B$ tokens:
\begin{equation}
\underbrace{\Scal_{q,h}^{(t)} \;=\; \underset{j}{\operatorname{Top-}k}\!\Big(\tfrac{1}{G}\!\sum_{g \in \mathrm{group}(h)}\! s_{g,q,j}^{(t)}\Big)}_{\text{per-token}}
\quad\;\;
\underbrace{\Scal_{b,h}^{(t)} \;=\; \underset{j}{\operatorname{Top-}k}\!\Big(\tfrac{1}{GB}\!\sum_{g \in \mathrm{group}(h)}\sum_{q \in b}\! s_{g,q,j}^{(t)}\Big)}_{\text{per-block}}
\label{eq:per-block-topk}
\end{equation}
where $j \in [N]$ indexes a prefix KV position ($N$ is the prefix length), and $s_{g,q,j}^{(t)}$ is determined by the estimation method: when $s$ is the exact attention score matrix, the resulting set is the \emph{oracle per-block KV subset} $\mathcal{S}_{b,h}^{*(t)}$; when a lightweight estimator like Quest substitutes a cheap approximation for $s$, the resulting set is the \emph{estimated per-block KV subset} $\hat{\mathcal{S}}_{b,h}^{(t)}$.

\begin{figure}[t!]
    \centering
    \includegraphics[width=\textwidth]{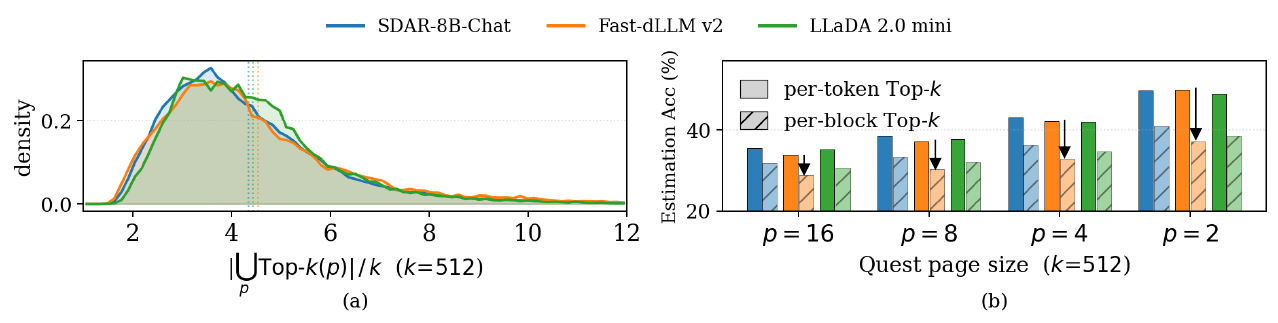}
    \caption{Sparse attention challenge on block dLLMs on LongBench, across three block-diffusion families. \emph{(a)} Distribution of $|\bigcup_p \mathrm{Top}\text{-}k(p)| / k$ at $k{=}512$, $t{=}1$. \emph{(b)} Quest~\citep{tang2024questqueryawaresparsityefficient} estimation accuracy at $k{=}512$, $t{=}1$, page sizes $p \in \{16, 8, 4, 2\}$.}
    \label{fig:challenge}
\end{figure}

\para{Estimation Accuracy Drop Under Per-Block Shared Selection}
A single per-block KV subset shared by $B$ tokens cannot satisfy each token's individual oracle simultaneously.
To show this empirically, we form the union of the $B$ per-token oracle top-$k$ subsets and measure its size as a multiple of $k$.
Fig.~\ref{fig:challenge}(a) reports this ratio at $k{=}512$ across three block-diffusion models on LongBench: the union averages $4.5\times$ the per-token budget, with a heavy tail beyond $10\times$.
Fig.~\ref{fig:challenge}(b) quantifies the resulting estimation accuracy drop using Quest~\citep{tang2024questqueryawaresparsityefficient}: across all three models and page sizes $p \in \{16, 8, 4, 2\}$, the per-block estimation accuracy falls below the per-token baseline by up to $25\%$.

\subsection{The Opportunity: Accurate Estimation via All-\texttt{[MASK]} Block Anchoring}
\label{subsec:allmask-knows}

While a single $k$-sized subset cannot satisfy every token simultaneously, the fact that 4.5$\times$ the per-token budget suffices to cover all $B$=32 tokens indicates that a substantial intersection of KV positions required across many tokens clearly exists.
Accurately identifying this intersection within the $k$-budget is therefore the central question for per-block sparse estimation under the constraint of \S\ref{subsec:challenge}.
We empirically discover that this intersection, the oracle per-block KV subset, barely drifts across the denoising trajectory, a phenomenon rooted in the block-diffusion training mechanism.
This stability allows us to invest one exact-attention pass at the first step to estimate the oracle accurately and reuse it across the remaining steps, amortizing the first-step cost across the trajectory and boosting estimation accuracy without sacrificing inference efficiency.

\begin{figure*}[t!]
    \centering
    \includegraphics[width=\textwidth]{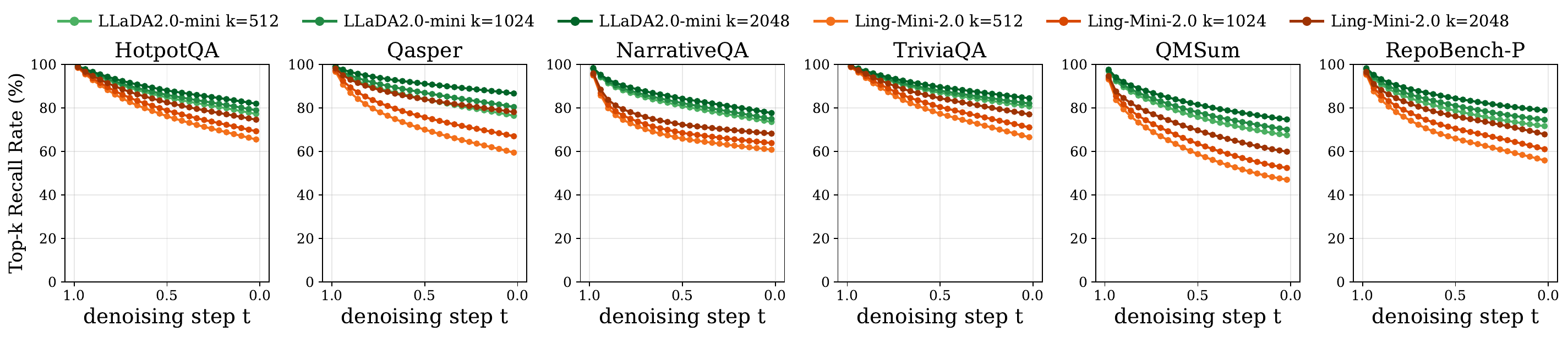}
    \caption{Per-block top-$k$ recall along the denoising trajectory on LongBench. \emph{Trained}: LLaDA2.0-mini; \emph{Backbone}: its pre-training checkpoint Ling-Mini-base-2.0. Recall is substantially higher after block-diffusion training across all tasks and $k$.}
    \label{fig:topk_consistency}
\end{figure*}

\para{Observation: Oracle Per-Block KV Subset Barely Drifts Across Denoising Steps}
We quantify this stability as the fraction of the all-\texttt{[MASK]} oracle ($t{=}1$) recovered at an arbitrary noise level $\tau \in [0,1)$:
\begin{equation}
r_{b,h}^{(\tau)} \;=\; \frac{\bigl|\Scal_{b,h}^{*(1)} \cap \Scal_{b,h}^{*(\tau)}\bigr|}{k},
\label{eq:topk-recall}
\end{equation}
LLaDA2.0-mini averages $85\%$ / $86\%$ / $89\%$ recall at $k{=}512$ / $1024$ / $2048$, uniform across all 6 LongBench tasks (Fig.~\ref{fig:topk_consistency}).
To check whether this anchoring is induced by block-diffusion training, we run the same measurement on its AR pre-training checkpoint (Ling-Mini-base-2.0): only $76\%$ / $79\%$ / $83\%$, pointing to training as the source.

\para{Mechanism: Anchoring as a Direct Consequence of the
Block-Diffusion Objective}
We trace this stability to the block-diffusion training procedure
itself: the prefix is kept as clean tokens while the current block
is corrupted at a random noise level, yielding a mask pattern
$\mathbf{m}$, and the model is trained for every $\mathbf{m}$ to
predict the ground-truth token at each masked position. Since the
prefix is identical, the only difference seen by a [MASK] token
under two mask patterns $\mathbf{m}^*$ and $\mathbf{m}$ is whether
each surrounding position is itself a [MASK] or a revealed token.
Yet training pushes the predictions under both patterns toward
the same ground-truth target, forcing the activations of
[MASK] and revealed positions to align across noise levels. This pressure
propagates from hidden states to queries and ultimately to the
per-block top-$k$ subset; we verify the stages (\textbf{S1}:
prediction $\to$ \textbf{S2}: hidden state $\to$ \textbf{S3}--
\textbf{S4}: query $\to$ \textbf{S5}: top-$k$) against AR
pre-training backbones in Table~\ref{tab:training_chain}.

\emph{(Stage 1) Cross-noise-level prediction agreement.}
The block-diffusion loss averages cross-entropy over the Bernoulli
mask schedule:
\begin{equation}
\mathcal{L}_{\text{block}}(\theta) \;=\; \mathbb{E}_{t \sim
\mathcal{U}(0,1),\;\mathbf{m} \sim \mathrm{Bern}(t)^B}\!\Big[\,
-\!\!\sum_{j:\,m_j=1}\! \log p_\theta\!\bigl(x_j \,\big|\,
\mathbf{m},\, \text{prefix}\bigr)\Big].
\label{eq:diffusion_loss}
\end{equation}
At each [MASK] token $j$, this is $D_{\mathrm{KL}}(\delta_{x_j} \,\|\,
p_\theta(\cdot|\mathbf{m},\text{prefix}))$, minimized for every
$\mathbf{m}$ independently — so predictions at different noise
levels collapse onto $\delta_{x_j}$ and thus onto each other
($D_{\mathrm{KL}}(p^{(\mathbf{m}^*)} \| p^{(\mathbf{m})})$ drops
43--63\%, col.~1).

\emph{(Stage 2) Hidden state alignment.}
With the prefix KV fixed, this pressure aligns the block's hidden
states across noise levels
($\|\tilde h^{(\mathbf{m}^*)} - \tilde h^{(\mathbf{m})}\|$ drops
8--19\%, col.~2).

\emph{(Stage 3) Per-position query gap.}
Hidden-state alignment propagates linearly to queries via
$q = W_Q h$. The per-position query gap
$A := \tfrac{1}{B}\sum_i \|\bar\delta_i\|$ with
$\bar\delta_i := \mathbb{E}_{\mathbf{m}}[q_i^{(\mathbf{m}^*)}
- q_i^{(\mathbf{m})}]$ shrinks 13--27\% (col.~3).

\emph{(Stage 4) Block-average query gap.}
Block averaging absorbs token-specific variance, keeping the gap
on $\bar q^{(t)} := \tfrac{1}{B}\sum_i q_i^{(t)}$ strictly below
the per-position gap (col.~4 < col.~3 in every family).
By Jensen's inequality,
\begin{equation*}
\big\|\mathbb{E}_{\mathbf{m}}[\bar q^{(\mathbf{m}^*)} - \bar q^{(\mathbf{m})}]\big\|
\;=\; \Big\|\tfrac{1}{B}\sum_{i=1}^{B} \bar\delta_i\Big\|
\;\le\; \tfrac{1}{B}\sum_{i=1}^{B} \|\bar\delta_i\| \;=\; A.
\end{equation*}
\emph{(Stage 5) Per-block top-$k$ stability.}
The stabilized block-average query, against the fixed prefix keys,
stabilizes the per-block top-$k$ subset (recall at $k{=}512$ rises
5--11 points, col.~5) — establishing $\mathbf{m}^*$ as a valid
anchor for the entire trajectory.
\begin{table}[t!]
\centering
\small
\setlength{\tabcolsep}{4pt}
\caption{Empirical verification of the five-stage alignment on
LongBench, comparing block-diffusion-trained models against their
AR pre-training backbones: cross-noise-level KL divergence of
predictions at masked positions (Stage~1), normalized L2 gap of
hidden states (Stage~2), per-position and block-average query L2
gaps (Stages~3--4), and top-$k$ recall (Stage~5). All L2 gaps
use unit-normalized vectors. Percentages are relative change
from backbone to trained.}
\label{tab:training_chain}
\resizebox{0.94\textwidth}{!}{%
\begin{tabular}{llccccc}
\toprule
& & {Logit} & {Hidden state} & \multicolumn{2}{c}{Query $\|\tilde q^{(\mathbf{m}^{\star})} - \tilde q^{(\mathbf{m})}\|$} & {Top-$k$ Recall} \\
& & {(Stage 1)} & {(Stage 2)} & \multicolumn{2}{c}{(Stages 3--4)} & {(Stage 5)} \\
\cmidrule(lr){3-3} \cmidrule(lr){4-4} \cmidrule(lr){5-6} \cmidrule(lr){7-7}
Family & Model & $D_{\mathrm{KL}}\!\bigl(p^{(\mathbf{m}^{\star})} \,\big\|\, p^{(\mathbf{m})}\bigr)$ $\downarrow$ & $\|\tilde h^{(\mathbf{m}^{\star})} - \tilde h^{(\mathbf{m})}\|$ $\downarrow$ & per-position $\downarrow$ & block-avg $\downarrow$ & $k{=}512$ $\uparrow$ \\
\midrule
\multirow{2}{*}{SDAR}
 & Qwen3-8B \emph{(backbone)}        & $2.40$              & $0.57$              & $0.38$              & $0.28$              & $0.808$              \\
 & SDAR-8B-Chat \emph{(trained)}     & $1.38\,(\!-\!43\%)$ & $0.53\,(\!-\!8\%)$  & $0.34\,(\!-\!13\%)$ & $0.22\,(\!-\!23\%)$ & $0.852\,(\!+\!5\%)$  \\
\midrule
\multirow{2}{*}{Fast-dLLM v2}
 & Qwen2.5-7B \emph{(backbone)}      & $5.66$              & $0.94$              & $0.38$              & $0.27$              & $0.753$              \\
 & Fast-dLLM v2 7B \emph{(trained)}  & $2.08\,(\!-\!63\%)$ & $0.78\,(\!-\!17\%)$ & $0.28\,(\!-\!27\%)$ & $0.18\,(\!-\!35\%)$ & $0.826\,(\!+\!10\%)$ \\
\midrule
\multirow{2}{*}{LLaDA2}
 & Ling-mini-base-2.0 \emph{(backbone)} & $2.61$              & $0.70$              & $0.45$              & $0.30$              & $0.769$              \\
 & LLaDA2.0-mini \emph{(trained)}       & $0.99\,(\!-\!62\%)$ & $0.57\,(\!-\!19\%)$ & $0.36\,(\!-\!20\%)$ & $0.24\,(\!-\!20\%)$ & $0.851\,(\!+\!11\%)$ \\
\bottomrule
\end{tabular}%
}
\end{table}

\begin{figure*}[t!]
    \centering
    \includegraphics[width=\textwidth]{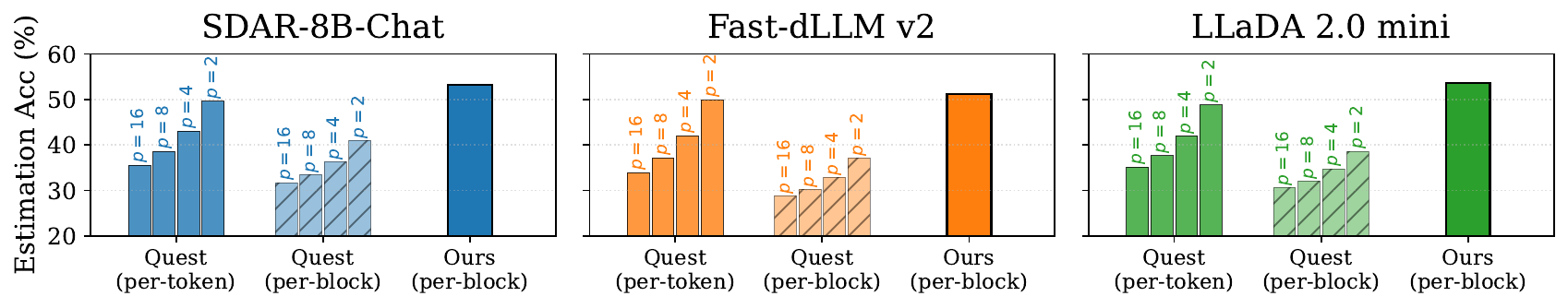}
    \caption{Estimation accuracy of different KV subset estimators at $k{=}512$ on LongBench, across three block-diffusion families.}
    \label{fig:opportunity}
\end{figure*}

\para{Our Proposal: One Exact Attention Pass at Step 1 Closes the Estimation Gap}
Concretely, we run one exact-attention pass on the all-\texttt{[MASK]} block at $t{=}1$, extract the exact per-block KV subset $\Scal_{b,h}^{*(1)}$, i.e., the oracle at the first step, and reuse it for all remaining steps (i.e., $\hat{\mathcal{S}}_{b,h}^{(t)} = \mathcal{S}_{b,h}^{*(1)}$ for all $t < 1$).
Because the per-block KV subset is computed exactly rather than approximated, our method removes the estimator-side error that lightweight methods like Quest accumulate on top of the per-block constraint.
As a result (Fig.~\ref{fig:opportunity}), our estimation accuracy surpasses Quest per-block at every page size, and matches or exceeds even Quest per-token at the finest setting $p{=}2$.

\section{\name: \texttt{[MASK]}-Guided Sparse Attention}
\label{sec:method}

\begin{algorithm}[t]
\caption{\name block inference}
\label{alg:mage}
\small
\begin{algorithmic}[0]
\Require Prefix KV cache $\mathcal{C}$, block size $B$, budget $k$, denoising steps $T$
\Ensure Generated block $\mathbf{x}$
\State Initialize block: $B$ \texttt{[MASK]} tokens
\Statex \textit{\textcolor{gray}{// Phase 1: exact selection at Step $1$ (All-\texttt{[MASK]}); two CUDA streams run concurrently}}
\end{algorithmic}
\vspace{2pt}
\noindent%
\begin{minipage}[t]{0.50\linewidth}
\begin{algorithmic}[0]
\Statex \textbf{\underline{Main Stream}}
\For{layer $\ell = 1, \ldots, L$}
    \State $(\mathbf{Q}^{(1)}_{\ell}, \mathbf{K}^{(1)}_{\ell}, \mathbf{V}^{(1)}_{\ell}) \gets \operatorname{QKVProj}_{\ell}\!\bigl(\mathbf{h}^{(1)}_{\ell-1}\bigr)$
    \State enqueue $\mathbf{Q}^{(1)}_{\ell}$ $\to$ \textsc{select} \Comment{non-blocking}
    \State $\mathbf{h}^{(1)}_{\ell} \gets \operatorname{Attn}_{\ell}\!\bigl(\mathbf{Q}^{(1)}_{\ell}, \mathbf{K}^{(1)}_{\ell}, \mathbf{V}^{(1)}_{\ell}, \mathcal{C}\bigr)$
    \State $\mathbf{h}^{(1)}_{\ell} \gets \operatorname{FFN}_{\ell}\!\bigl(\mathbf{h}^{(1)}_{\ell}\bigr)$
\EndFor
\end{algorithmic}
\end{minipage}%
\hfill\vline\hfill%
\begin{minipage}[t]{0.46\linewidth}
\begin{algorithmic}[0]
\Statex \textbf{\underline{Select Stream}}
\For{layer $\ell = 1, \ldots, L$}
    \State wait for $\mathbf{Q}^{(1)}_{\ell}$
    \State $\mathbf{A}^{(1)}_{\ell} \gets \operatorname{softmax}\!\bigl(\mathbf{Q}^{(1)}_{\ell}\mathbf{K}^{\top}/\sqrt{d}\bigr)$
    \For{each KV head $h$}
        \State $\bar{\alpha}_{\ell,h} \gets \tfrac{1}{GB}\!\sum_{g,q} \mathbf{A}^{(1)}_{\ell}[g, q, :]$
        \State $\mathcal{S}^{*\ell}_{b,h} \gets \operatorname{Top-}k(\bar{\alpha}_{\ell,h})$
    \EndFor
\EndFor
\end{algorithmic}
\end{minipage}
\vspace{2pt}
\begin{algorithmic}[0]
\State \textbf{synchronize}(\textsc{main}, \textsc{select})\hfill\textit{\textcolor{gray}{// Phase 1 complete; all $\mathcal{S}^{*\ell}_{b,h}$ ready}}
\Statex \textit{\textcolor{gray}{// Phase 2: sparse denoising for Step $s = 2, \ldots, T$}}
\For{$s = 2$ to $T$}
    \For{each layer $\ell$}
        \State $(\mathbf{Q}_{\ell}, \mathbf{K}_{\ell}, \mathbf{V}_{\ell}) \gets \operatorname{QKVProj}_{\ell}\!\bigl(\mathbf{h}_{\ell-1}\bigr)$
        \For{each KV head $h$}
            \State $\mathbf{o}_{\ell,h} \gets \operatorname{SparseAttn}\!\bigl(\mathbf{Q}_{\ell,h}, \mathbf{K}_{\ell,h}, \mathbf{V}_{\ell,h}, \mathcal{C}[\mathcal{S}^{*\ell}_{b,h}]\bigr)$
        \EndFor
        \State $\mathbf{h}_{\ell} \gets \operatorname{FFN}_{\ell}\!\bigl(\mathbf{o}_{\ell}\bigr)$
    \EndFor
    \State Unmask high-confidence positions in the block
\EndFor
\State $\mathcal{C} \gets \mathcal{C} \,\Vert\, \mathbf{KV}(\mathbf{x})$
\State \Return $\mathbf{x}$
\end{algorithmic}
\end{algorithm}

We present \name (\texttt{[MASK]}-Guided Sparse Attention), the first sparse attention method tailored to block-diffusion LLMs (Fig.~\ref{fig:overview} and Algorithm~\ref{alg:mage}). Building on the one-shot All-\texttt{[MASK]} selection of \S\ref{subsec:allmask-knows}, \name introduces an asynchronous dual-stream execution that hides the selection cost behind the layer's FFN, leaving the main forward pass as \name's sole non-sparse work. The procedure is training-free and applies to any pretrained block-diffusion LLM.

\para{Phase 1: Exact Selection at Step 1}
At the start of each block's trajectory the $B$ block positions are all \texttt{[MASK]}, and Phase 1 splits work across two CUDA streams (Algorithm~\ref{alg:mage}): a \textsc{main} stream runs a full forward pass (FlashAttention~\citep{dao2022flashattentionfastmemoryefficientexact} $+$ FFN) without materializing the score matrix, while a \textsc{select} stream independently re-derives the prefix attention scores. At each layer $\ell$, the \textsc{main} stream hands off $\mathbf{Q}^{(1)}_\ell$ right after the QKV projection and proceeds to $\operatorname{FlashAttn}_\ell$ and $\operatorname{FFN}_\ell$; the \textsc{select} stream computes $\mathbf{A}^{(1)}_\ell = \operatorname{softmax}(\mathbf{Q}^{(1)}_\ell \mathbf{K}^{\top}/\sqrt{d}) \in \mathbb{R}^{G \times B \times N}$ ($N$ the prefix length, $G$ the query heads per KV group) and, for each KV head $h$, averages over the $GB$ queries that share the head to select the top-$k$ prefix positions:
\begin{equation}
\mathcal{S}^{*\ell}_{b,h} \;=\; \operatorname{Top-}k\!\Big(\,\tfrac{1}{GB}\!\sum_{g \in \mathrm{group}(h)} \sum_{q \in b}\, \mathbf{A}^{(1)}_{\ell}[g, q, :]\,\Big).
\label{eq:mage-selection}
\end{equation}
This is the per-block KV subset of Eq.~\ref{eq:per-block-topk} computed from the exact attention scores at Step~1, so no estimator approximation is involved. The dual-stream design hides the \textsc{select} stream's cost behind $\operatorname{FFN}_\ell$: $\operatorname{FFN}_\ell$ is compute-intensive while $\mathbf{Q}\mathbf{K}^{\top}$ is bound by HBM bandwidth, so the two kernels target different hardware resources and run without contention. The streams synchronize once at the end of Phase 1, leaving the main forward pass as \name's sole non-sparse cost---amortized to $1/T$ per step over the $T$-step trajectory.

\para{Phase 2: Sparse Denoising for Step $s=2,...,T$}
At every subsequent denoising step, attention at each layer/head retrieves only the $k$ KV pairs at indices $\mathcal{S}^{*\ell}_{b,h}$ and ignores the remaining prefix entries. The model then unmasks high-confidence positions in the block as in standard block-diffusion decoding. Because the selection is fixed once at Step $1$ and reused unchanged across all heads, layers, and steps within the block, no re-estimation is performed at any $s \geq 2$.

\section{Evaluation}
\label{sec:experiments}

\paragraph{Setup.}
We evaluate \name on three open block-diffusion LLM families: Fast-dLLM v2 7B~\citep{wu2025fastdllmv2efficientblockdiffusion}, SDAR-8B-Chat~\citep{cheng2025sdarsynergisticdiffusionautoregressionparadigm}, and LLaDA 2.0 mini~\citep{bie2025llada20scalingdiffusionlanguage}. We compare against the dense baseline---\emph{Exact Attention} (FlashInfer~\citep{ye2025flashinferefficientcustomizableattention})---and two sparse-attention baselines, Quest~\citep{tang2024questqueryawaresparsityefficient} and SparseD~\citep{wang2025sparsedsparseattentiondiffusion}, both adapted to block-diffusion decoding via per-block KV selection. Following their original designs, Quest re-estimates the per-block top-$k$ at every denoising step with page size $p{=}16$, and SparseD performs exact attention for the first 20\% of the total denoising steps across all blocks before reusing the captured sparse pattern for the remaining 80\%. All sparse methods share the same KV budget $k \in \{256, 512, 1024\}$ and retain exact attention for the first two layers (layers 1--2) across all methods for a fair comparison. We use block size $B{=}32$ and unmasking confidence threshold $p_{\text{conf}}{=}0.95$~\citep{bie2025llada20scalingdiffusionlanguage}. All sparse-attention kernels are implemented with FlashInfer, and experiments run on a NVIDIA H100 GPU.

\subsection{Accuracy}
\label{subsec:accuracy}

\begin{table}[t]
\centering
\scriptsize
\setlength{\tabcolsep}{3pt}
\caption{LongBench per-task and average accuracy for all (model, method, budget) configurations.}
\label{tab:acc_sparsed}
\resizebox{0.94\textwidth}{!}{%
\begin{tabular}{ll|c|ccc|ccc|ccc}
\toprule
 &  & & \multicolumn{3}{c|}{$k{=}256$} & \multicolumn{3}{c|}{$k{=}512$} & \multicolumn{3}{c}{$k{=}1024$} \\
\cmidrule(lr){4-6} \cmidrule(lr){7-9} \cmidrule(lr){10-12}
Model & Task ($\uparrow$) & Exact & MAGE & Quest  & SparseD & MAGE & Quest  & SparseD  & MAGE & Quest & SparseD \\
\midrule
\multirow{7}{*}{Fast-dLLM v2 7B} & HotpotQA & 42.38 & 37.98 & 32.13 & \textbf{38.72} & 39.33 & 34.74 & \textbf{40.93} & 41.29 & 38.06 & \textbf{42.32} \\
 & NrtvQA & 23.63 & \textbf{22.56} & 14.98 & 21.83 & \textbf{24.23} & 15.31 & 22.93 & 22.42 & 19.06 & \textbf{23.42} \\
 & Qasper & 36.66 & \textbf{31.43} & 25.82 & 25.05 & \textbf{33.41} & 31.60 & 31.65 & 34.05 & \textbf{34.54} & 33.50 \\
 & QMSum & 24.16 & \textbf{23.92} & 21.68 & 20.95 & \textbf{23.94} & 22.85 & 22.90 & \textbf{24.79} & 23.10 & 23.34 \\
 & RepoB-P & 51.84 & 49.81 & 41.20 & \textbf{51.05} & 51.71 & 46.69 & \textbf{51.96} & \textbf{52.36} & 49.57 & 51.88 \\
 & TriviaQA & 75.49 & \textbf{75.17} & 57.72 & 69.66 & \textbf{75.65} & 61.86 & 70.20 & \textbf{74.17} & 66.24 & 72.51 \\
\cmidrule(lr){2-12}
 & Avg & 42.36 & \textbf{40.15} & 32.26 & 37.88 & \textbf{41.38} & 35.51 & 40.09 & \textbf{41.51} & 38.43 & 41.16 \\
\midrule
\multirow{7}{*}{SDAR-8B-Chat} & HotpotQA & 59.59 & \textbf{55.87} & 45.81 & 50.74 & \textbf{57.17} & 49.93 & 56.54 & 57.00 & 55.66 & \textbf{60.03} \\
 & NrtvQA & 28.94 & \textbf{28.86} & 20.11 & 23.01 & \textbf{28.74} & 24.19 & 26.22 & \textbf{29.31} & 26.80 & 28.80 \\
 & Qasper & 31.02 & \textbf{30.89} & 22.70 & 17.98 & \textbf{31.11} & 25.68 & 26.47 & \textbf{31.51} & 28.19 & 29.47 \\
 & QMSum & 24.41 & \textbf{23.57} & 18.90 & 20.30 & \textbf{23.97} & 21.48 & 21.75 & \textbf{23.59} & 23.27 & 23.00 \\
 & RepoB-P & 50.04 & 46.56 & 33.17 & \textbf{46.66} & \textbf{48.41} & 41.29 & 48.38 & 48.66 & 46.78 & \textbf{48.95} \\
 & TriviaQA & 88.83 & \textbf{89.63} & 78.32 & 78.75 & \textbf{88.21} & 86.23 & 86.70 & 88.16 & \textbf{89.36} & 87.39 \\
\cmidrule(lr){2-12}
 & Avg & 47.14 & \textbf{45.90} & 36.50 & 39.57 & \textbf{46.27} & 41.47 & 44.34 & \textbf{46.37} & 45.01 & 46.27 \\
\midrule
\multirow{7}{*}{LLaDA 2.0 mini} & HotpotQA & 60.29 & \textbf{56.75} & 48.56 & 54.20 & 58.48 & 55.37 & \textbf{59.65} & 59.61 & 57.39 & \textbf{60.15} \\
 & NrtvQA & 27.91 & \textbf{26.19} & 13.74 & 21.95 & \textbf{28.48} & 18.40 & 24.90 & \textbf{28.16} & 24.30 & 26.22 \\
 & Qasper & 46.19 & \textbf{45.23} & 32.89 & 32.87 & \textbf{46.41} & 40.50 & 44.25 & \textbf{46.38} & 45.65 & 44.72 \\
 & QMSum & 22.86 & \textbf{22.82} & 19.42 & 20.21 & \textbf{22.86} & 21.43 & 22.05 & \textbf{23.06} & 22.02 & 22.32 \\
 & RepoB-P & 57.11 & 55.52 & 46.86 & \textbf{55.88} & 56.18 & 53.83 & \textbf{56.74} & 56.71 & \textbf{57.59} & 57.33 \\
 & TriviaQA & 86.43 & \textbf{87.49} & 80.71 & 85.15 & \textbf{86.70} & 84.26 & 86.43 & 86.42 & \textbf{86.93} & 86.13 \\
\cmidrule(lr){2-12}
 & Avg & 50.13 & \textbf{49.00} & 40.36 & 45.04 & \textbf{49.85} & 45.63 & 49.00 & \textbf{50.06} & 48.98 & 49.48 \\
\bottomrule
\end{tabular}
}
\end{table}

\paragraph{LongBench.}
We evaluate on six LongBench~\citep{bai2024longbenchbilingualmultitaskbenchmark} long-context tasks spanning single-document QA (NarrativeQA, Qasper), multi-document QA (HotpotQA), few-shot learning (TriviaQA), summarization (QMSum), and code completion (RepoBench-P). Table~\ref{tab:acc_sparsed} reports per-task scores (F1 / ROUGE-L / EditSim depending on task) and averages across the six tasks for each (model, method, budget) configuration.

At $k{=}512$, \name reaches near-lossless accuracy with average drops of only $0.98$/$0.87$/$0.28$ on Fast-dLLM v2 7B/SDAR-8B-Chat/LLaDA 2.0 mini, far below Quest ($6.85$/$5.67$/$4.50$) and SparseD ($2.27$/$2.80$/$1.13$); even at $k{=}256$, \name's drops ($2.21$/$1.24$/$1.13$) stay well under Quest's ($10.10$/$10.64$/$9.77$) and SparseD's ($4.48$/$7.57$/$5.09$).
\name's margin is largest on retrieval-heavy tasks (NarrativeQA, HotpotQA), because \name accurately extracts the KV subset each block needs whereas Quest's page-anchoring incurs approximation error and SparseD never refreshes its captured pattern, propagating errors across later blocks.

\begin{figure}[t]
    \centering
    \includegraphics[width=\textwidth]{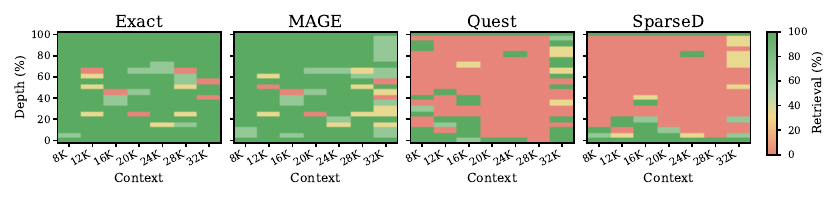}
    \caption{NIAH retrieval on LLaDA 2.0 mini at $k{=}256$. $x$: context length (8K--32K); $y$: needle depth (0--100\%); color: retrieval score. \name $\approx$ Exact, SparseD collapses on most cells, Quest fails across most cells beyond 8K. B=32}
    \label{fig:niah}
\end{figure}

\paragraph{Needle-in-a-Haystack.}
Fig.~\ref{fig:niah} shows NIAH heatmaps for LLaDA 2.0 mini at $k{=}256$ over context lengths 8K--32K. \name closely tracks Exact with only a few scattered failures, whereas SparseD shows widespread red/yellow cells across most (context, depth) positions, and Quest collapses into near-total failure beyond 12K. At 8K, retrieval scores are \name~97, SparseD~17, Quest~41 (Exact~98); per-context scores for all three models at $k \in \{256,512,1024\}$ are in Table~\ref{tab:niah} (Appendix~\ref{app:niah}).

\subsection{Efficiency}
\label{subsec:efficiency}

\para{End-to-End Speedup}
\begin{figure*}[t]
    \centering
    \includegraphics[width=\textwidth]{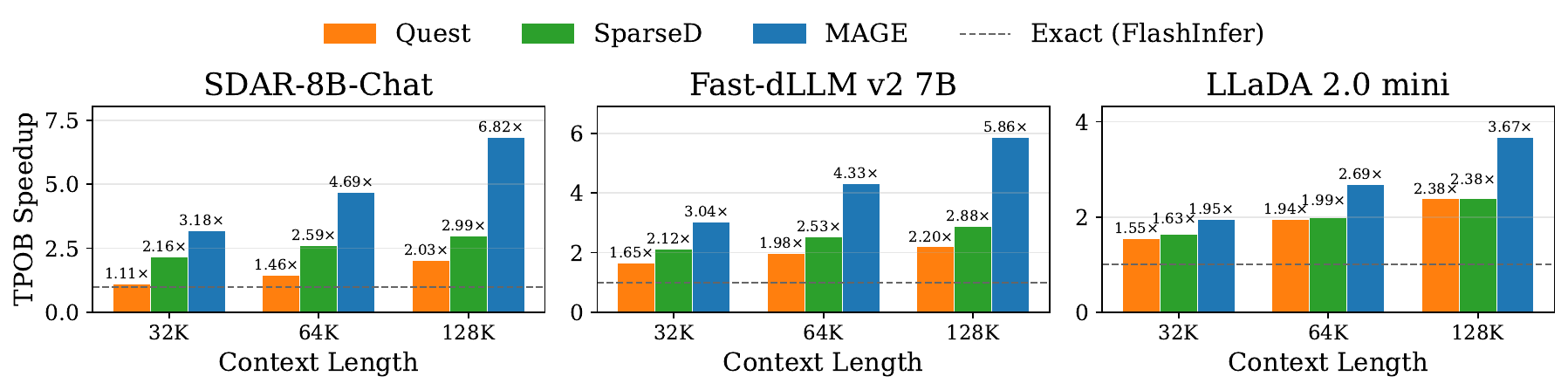}
    \caption{End-to-end TPOB (time-per-output-block) speedup over Exact Attention (FlashInfer) on three block-diffusion LLM families. Budget $k{=}1024$, $32$ steps/block.}
    \label{fig:end2end}
\end{figure*}
Fig.~\ref{fig:end2end} reports \name's end-to-end TPOB (time-per-output-block) speedup over Exact Attention (FlashInfer) on all three block-diffusion families at three context lengths ($32$K, $64$K, $128$K) with a fixed sparse budget $k{=}1024$ and $n{=}32$ block steps; the $64$K and $128$K settings use NTK-based RoPE extrapolation~\citep{liu2025longlladaunlockinglongcontext} to extend each model's context window. Across every (model, context) pair, the speedup grows with context length as the savings of sparse attention scale with $N$. At $128$K context, \name's speedup over Exact reaches $\mathbf{6.82}\times$ on SDAR-8B-Chat, $\mathbf{5.86}\times$ on Fast-dLLM v2 7B, and $\mathbf{3.67}\times$ on LLaDA 2.0 mini --- consistently ahead of both Quest and SparseD on every (model, context) cell, running up to $\mathbf{3.35}\times$ faster than Quest and $\mathbf{2.28}\times$ faster than SparseD. That \name dominates across three independent block-diffusion implementations indicates the gain is robust to the model rather than a property of any single one. The same ranking holds under shorter block-step budgets $n \in \{24, 16, 8\}$; see Appendix~\ref{app:end2end_n}.

\para{Latency Breakdown}
\begin{figure*}[t]
    \centering
    \includegraphics[width=\textwidth]{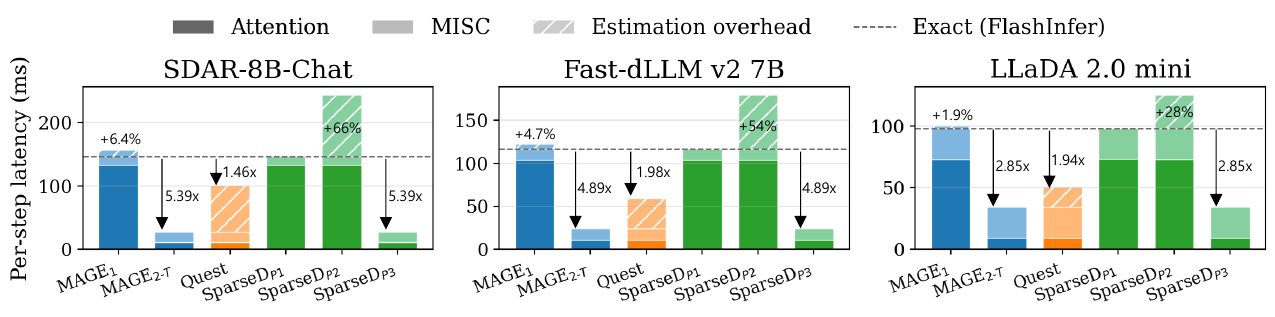}
    \caption{Per-step latency breakdown at $64$K, $k{=}1024$. Bars decompose one denoising step into Attention, Estimation overhead (hatched), and MISC; the dashed line marks Exact Attention (FlashInfer).}
    \label{fig:breakdown}
\end{figure*}
Fig.~\ref{fig:breakdown} decomposes latency of one denoising step at $64$K, $k{=}1024$ into Attention, Estimation overhead, and MISC, with Exact as a dashed reference. \name's first step (\name$_1$) adds top-$k$ selection on top of an exact pass, but the selection runs on the asynchronous \textsc{select} stream in parallel with FFN (Algorithm~\ref{alg:mage}), keeping critical-path overhead at $\mathbf{+1.9\%}$--$\mathbf{+6.4\%}$ over Exact. The remaining $n{-}1$ steps (\name$_{2\text{-}T}$) run pure sparse attention at $\mathbf{2.85}\times$--$\mathbf{5.39}\times$ over Exact, amortizing the single exact pass per block. Quest recomputes its page-anchor estimate on the critical path at every step, capping per-step speedup at $\mathbf{1.46}\times$--$\mathbf{1.98}\times$. SparseD runs exact attention for the first 20\% of steps across the trajectory (P1), takes a capture step at the boundary (P2, $\mathbf{+28\%}$--$\mathbf{+66\%}$), and runs sparse attention for the remaining 80\% (P3, matching \name$_{2\text{-}T}$); \name keeps only $1/n$ of steps exact while SparseD keeps 20\%, so the gap accumulates end-to-end.

\section{Limitations}
MAGE targets the attention bottleneck that dominates long-context inference, so its end-to-end benefit diminishes in short-context regimes where FFN and other non-attention computations account for most of the latency. In addition, MAGE reduces KV cache access time but not its memory footprint; alleviating the memory pressure of long-context KV caches requires complementary techniques such as KV offloading~\citep{lee2024infinigenefficientgenerativeinference, sun2025shadowkvkvcacheshadows, kwon2026heraldhighthroughputblockdiffusion} or cache compression, which we leave to future work.

\section{Conclusion}
\label{sec:conclusion}

We present \name, a training-free sparse-attention method for block-diffusion LLMs. By exploiting the All-\texttt{[MASK]} anchoring property induced by block-diffusion training, \name runs one exact attention pass at Step $1$ and reuses the resulting per-block top-$k$ indices across the remaining $T{-}1$ steps, with selection cost hidden asynchronously behind the FFN. Across three block-diffusion families, \name matches Exact Attention at $k{=}512$ with near-lossless accuracy while achieving up to $6.82\times$ end-to-end speedup at $128$K context.

\section{Acknowledgments}
This work was partly supported by Institute of Information \& communications Technology Planning \& Evaluation (IITP) grant funded by the Korea government (MSIT) (No. RS-2026-25524173, Ultra-Long-Term Hierarchical Memory and Reasoning Architecture for Next-Generation Omnimodal Agents).
Additional support was provided by Institute of Information \& communications Technology Planning \& Evaluation (IITP) under the Graduate School of Artificial Intelligence Semiconductor (IITP-2026-RS-2023-00256081) grant funded by the Korea government (MSIT).
Samsung Electronics Co., Ltd also supported this work (IO260225-15628-01).

\newpage
\bibliographystyle{abbrvnat}
\bibliography{references}

\newpage
\appendix
\section{NIAH Scores Across Models}
\label{app:niah}

Table~\ref{tab:niah} reports per-context-length NIAH retrieval scores at 8K
and 32K context for Exact Attention and the three sparse methods (\name,
Quest, SparseD) at $k \in \{256, 512, 1024\}$ across all three
block-diffusion model families. Each cell is the mean retrieval score
over the 21 needle-depth grid at that context length. Bold marks the
best sparse method (\name / Quest / SparseD) per column within each
model. Fig.~\ref{fig:niah} in the main text visualizes the
underlying (context, depth) cells for LLaDA 2.0 mini at $k{=}256$.

\begin{table*}[h!]
\centering
\footnotesize
\setlength{\tabcolsep}{2.5pt}
\caption{Needle-in-a-Haystack accuracy (\%) at 8K and 32K context, averaged over depth positions, for sparse budgets $k\in\{256,512,1024\}$ across three block-diffusion models. Bold = best sparse method (MAGE / Quest / SparseD) per column within each model.}
\label{tab:niah}
\resizebox{\textwidth}{!}{%
\begin{tabular}{l|cccccc|cccccc|cccccc}
\toprule
 & \multicolumn{6}{c|}{\textit{SDAR-8B-Chat}} & \multicolumn{6}{c|}{\textit{LLaDA 2.0 mini}} & \multicolumn{6}{c}{\textit{Fast-dLLM v2 7B}} \\
 & \multicolumn{3}{c}{8K} & \multicolumn{3}{c|}{32K} & \multicolumn{3}{c}{8K} & \multicolumn{3}{c|}{32K} & \multicolumn{3}{c}{8K} & \multicolumn{3}{c}{32K} \\
Method & 256 & 512 & 1024 & 256 & 512 & 1024 & 256 & 512 & 1024 & 256 & 512 & 1024 & 256 & 512 & 1024 & 256 & 512 & 1024 \\
\midrule
Exact & 100 & 100 & 100 & 57 & 57 & 57 & 98 & 98 & 98 & 90 & 90 & 90 & 100 & 100 & 100 & 46 & 46 & 46 \\
\name & \textbf{100} & \textbf{100} & \textbf{100} & \textbf{76} & \textbf{71} & \textbf{67} & \textbf{97} & \textbf{98} & 98 & \textbf{62} & 68 & \textbf{76} & 89 & 94 & \textbf{100} & \textbf{33} & \textbf{37} & 46 \\
Quest & 90 & 100 & 100 & 16 & 16 & 16 & 41 & 48 & 60 & 57 & \textbf{70} & 75 & \textbf{97} & 98 & 100 & 22 & 35 & \textbf{52} \\
SparseD & 68 & 100 & 100 & 10 & 38 & 49 & 17 & 68 & \textbf{100} & 25 & 65 & 70 & 56 & \textbf{100} & 100 & 13 & 27 & 48 \\
\bottomrule
\end{tabular}
}
\end{table*}

\section{End-to-end Speedup at Shorter Block-step Budgets}
\label{app:end2end_n}

\begin{figure*}[h!]
    \centering
    \includegraphics[width=\textwidth]{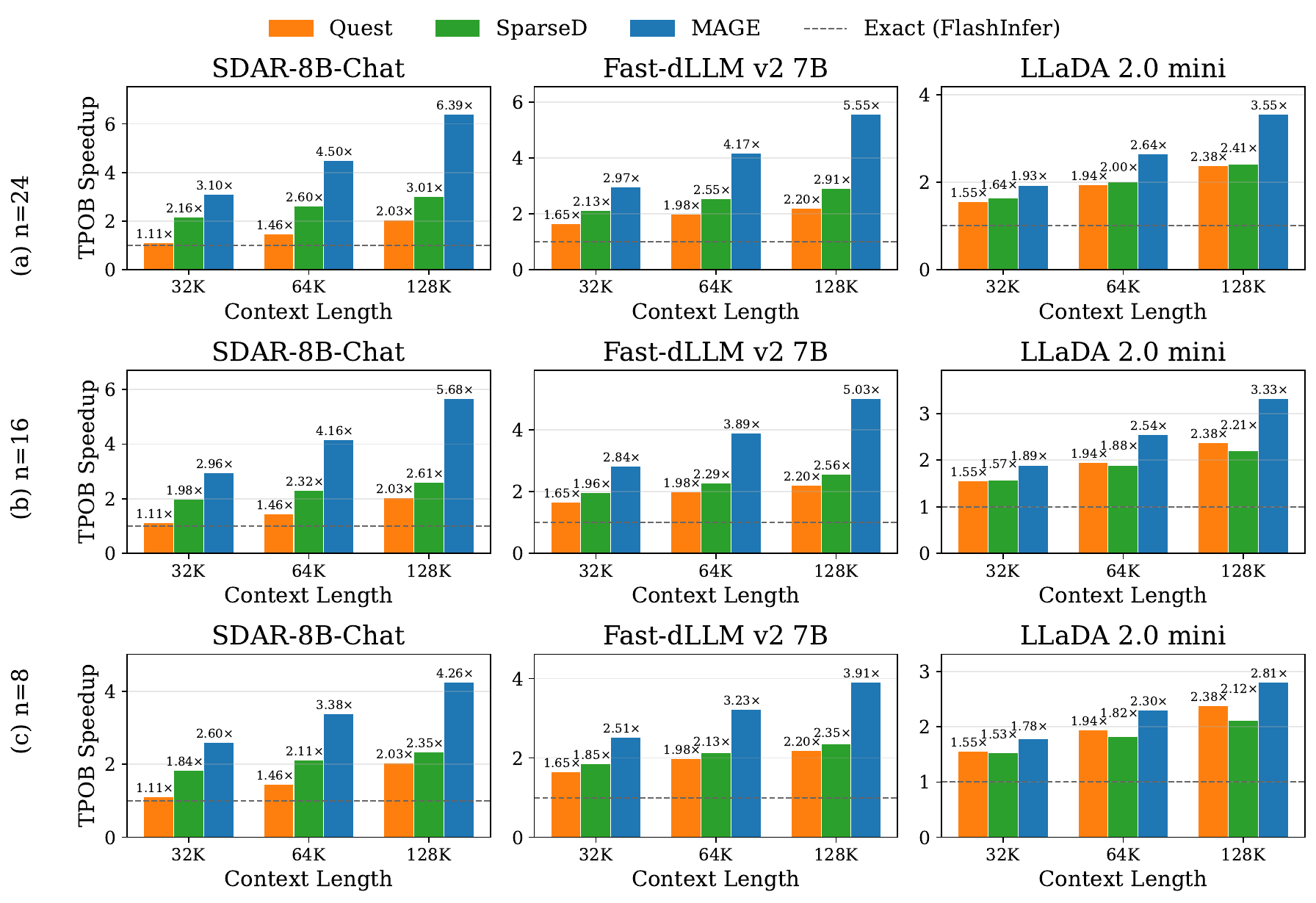}
    \caption{End-to-end TPOB speedup over Exact Attention (FlashInfer) at (a) $n{=}24$ steps/block, (b) $n{=}16$ steps/block, (c) $n{=} 8$ steps/block.  budget $k{=}1024$.}
    \label{fig:end2end_ablation}
\end{figure*}

The headline speedups in Fig.~\ref{fig:end2end} use $n{=}32$ steps/block; Fig.~\ref{fig:end2end_ablation} extends this to shorter budgets ($n{=}24, 16, 8$) on the same three model families. \name remains the fastest method on every (model, context) cell, with the absolute speedup tightening slightly at $n{=}8$ as its first-step dense pass amortizes over fewer sparse steps.

\section{Per-Token vs.\ Per-Block Sparse Attention Kernels}
\label{app:per-token}
A block could in principle let each of its $B$ tokens attend to its own per-token KV subset. We implement this alternative by treating each token as an independent decode request with its own KV-cache accesses, executed with FlashInfer's batched paged-attention kernel, and compare it with the per-block kernel of \name, in which all $B$ queries attend to a single shared KV subset (Table~\ref{tab:per-token}). The per-block kernel is $2.5$--$7.9\times$ faster, because the irregular accesses of the per-token kernel offset most of the benefit of reading fewer KV entries. A shared per-block KV subset is therefore necessary for efficient sparse attention in block dLLMs.
\begin{table}[h]
\centering
\caption{Sparse attention kernel latency with per-block and per-token KV subsets ($B{=}32$, $k{=}1024$, H100).}
\label{tab:per-token}
\small
\begin{tabular}{rrrr}
\toprule
Request batch & Per-block (ms) & Per-token (ms) & Per-block speedup \\
\midrule
1  & 0.026 & 0.066 & $2.5\times$ \\
4  & 0.047 & 0.242 & $5.2\times$ \\
16 & 0.110 & 0.874 & $7.9\times$ \\
64 & 0.469 & 3.711 & $7.9\times$ \\
\bottomrule
\end{tabular}
\end{table}

\newpage
\section*{NeurIPS Paper Checklist}

\begin{enumerate}

\item {\bf Claims}
    \item[] Question: Do the main claims made in the abstract and introduction accurately reflect the paper's contributions and scope?
    \item[] Answer: \answerYes{}
    \item[] Justification: The abstract and introduction accurately state our identification of the per-block selection challenge, the All-[MASK] anchoring property, and MAGE's training-free design.

\item {\bf Limitations}
    \item[] Question: Does the paper discuss the limitations of the work performed by the authors?
    \item[] Answer: \answerYes{}
    \item[] Justification: We discuss limitations in a dedicated Limitations section, noting that MAGE's benefit is concentrated in long-context regimes and that it reduces KV cache access time but not its memory footprint.

\item {\bf Theory assumptions and proofs}
    \item[] Question: For each theoretical result, does the paper provide the full set of assumptions and a complete (and correct) proof?
    \item[] Answer: \answerYes{}
    \item[] Justification: The derivation of the block-average query gap bound (Eqs. 4-7) states its assumptions and is presented in full in §3.2.

    \item {\bf Experimental result reproducibility}
    \item[] Question: Does the paper fully disclose all the information needed to reproduce the main experimental results of the paper to the extent that it affects the main claims and/or conclusions of the paper (regardless of whether the code and data are provided or not)?
    \item[] Answer: \answerYes{}
    \item[] Justification: MAGE is a training-free algorithm fully specified in Algorithm 1, with all experimental settings (models, baselines, budgets, hyperparameters, hardware) detailed in §5.

\item {\bf Open access to data and code}
    \item[] Question: Does the paper provide open access to the data and code, with sufficient instructions to faithfully reproduce the main experimental results, as described in supplemental material?
    \item[] Answer: \answerYes{}
    \item[] Justification: We provide an anonymized code package as supplemental material, including scripts to reproduce the main experiments on the publicly available models and benchmarks used in §5.

\item {\bf Experimental setting/details}
    \item[] Question: Does the paper specify all the training and test details (e.g., data splits, hyperparameters, how they were chosen, type of optimizer) necessary to understand the results?
    \item[] Answer: \answerYes{}
    \item[] Justification: §5 specifies all evaluation settings including models, baselines, sparse budgets, block size, decoding threshold, and hardware.

\item {\bf Experiment statistical significance}
    \item[] Question: Does the paper report error bars suitably and correctly defined or other appropriate information about the statistical significance of the experiments?
    \item[] Answer: \answerNo{}
    \item[] Justification: Speed measurements are reported as the median over 30 runs to mitigate timing noise, though error bars are not included; accuracy results follow the standard deterministic evaluation protocol of LongBench and NIAH.

\item {\bf Experiments compute resources}
    \item[] Question: For each experiment, does the paper provide sufficient information on the computer resources (type of compute workers, memory, time of execution) needed to reproduce the experiments?
    \item[] Answer:  \answerYes{}
    \item[] Justification: §5 specifies that all experiments run on a single NVIDIA H100 GPU.

\item {\bf Code of ethics}
    \item[] Question: Does the research conducted in the paper conform, in every respect, with the NeurIPS Code of Ethics \url{https://neurips.cc/public/EthicsGuidelines}?
    \item[] Answer: \answerYes{}
    \item[] Justification:  We have reviewed the NeurIPS Code of Ethics and our work conforms to it.

\item {\bf Broader impacts}
    \item[] Question: Does the paper discuss both potential positive societal impacts and negative societal impacts of the work performed?
    \item[] Answer: \answerNA{}{}
    \item[] Justification: MAGE is a general-purpose inference acceleration technique with no direct path to specific societal applications beyond those of the underlying LLMs it accelerates.

\item {\bf Safeguards}
    \item[] Question: Does the paper describe safeguards that have been put in place for responsible release of data or models that have a high risk for misuse (e.g., pre-trained language models, image generators, or scraped datasets)?
    \item[] Answer: \answerNA{}
    \item[] Justification: We release no new models or datasets; MAGE is an inference-time technique applied to existing publicly available models.

\item {\bf Licenses for existing assets}
    \item[] Question: Are the creators or original owners of assets (e.g., code, data, models), used in the paper, properly credited and are the license and terms of use explicitly mentioned and properly respected?
    \item[] Answer: \answerYes{}
    \item[] Justification:  All models, benchmarks, kernels, and baseline methods used are publicly available and properly cited in the paper.

\item {\bf New assets}
    \item[] Question: Are new assets introduced in the paper well documented and is the documentation provided alongside the assets?
    \item[] Answer: \answerYes{}
    \item[] Justification: The supplemental code package includes documentation describing how to run MAGE and reproduce the experiments in §5.

\item {\bf Crowdsourcing and research with human subjects}
    \item[] Question: For crowdsourcing experiments and research with human subjects, does the paper include the full text of instructions given to participants and screenshots, if applicable, as well as details about compensation (if any)?
    \item[] Answer: \answerNA{}
    \item[] Justification: This work does not involve crowdsourcing or human subjects.

\item {\bf Institutional review board (IRB) approvals or equivalent for research with human subjects}
    \item[] Question: Does the paper describe potential risks incurred by study participants, whether such risks were disclosed to the subjects, and whether Institutional Review Board (IRB) approvals (or an equivalent approval/review based on the requirements of your country or institution) were obtained?
    \item[] Answer: \answerNA{}
    \item[] Justification: This work does not involve human subjects research.

\item {\bf Declaration of LLM usage}
    \item[] Question: Does the paper describe the usage of LLMs if it is an important, original, or non-standard component of the core methods in this research? Note that if the LLM is used only for writing, editing, or formatting purposes and does \emph{not} impact the core methodology, scientific rigor, or originality of the research, declaration is not required.
    \item[] Answer: \answerNA{}
    \item[] Justification: LLMs are the target of our acceleration method but are not used as a component of the core methodology.

\end{enumerate}

\end{document}